%% file: author.tex
\documentclass{svproc}
\usepackage{url}

\usepackage{newtxtext}       %
\usepackage[varvw]{newtxmath}       

\usepackage[font=small,labelfont=bf]{caption}
\usepackage{subcaption}

\usepackage[export]{adjustbox}
\usepackage{booktabs}
\usepackage[utf8]{inputenc}
\usepackage{pgfplots}
\usepackage{tikz,tikzscale}
\usepackage{adjustbox}
\usepackage{svg}
\usepackage{pifont}
\usepackage{footmisc}
\usepackage{hyperref}

\newcommand{\R}{\mathbb{R}}

\DeclareUnicodeCharacter{2212}{−}
\usepgfplotslibrary{groupplots,dateplot}
\usetikzlibrary{patterns,shapes.arrows}
\pgfplotsset{compat=newest}

\newcommand{\cmark}{\ding{51}}%
\newcommand{\xmark}{\ding{55}}%

\begin{document}
\mainmatter              
\title{VMAS: A Vectorized Multi-Agent Simulator for Collective Robot Learning}
\titlerunning{VMAS: A Vectorized Multi-Agent Simulator for Collective Robot Learning}  
%
\author{Matteo Bettini \and Ryan Kortvelesy \and Jan Blumenkamp \and Amanda Prorok}
\authorrunning{Matteo Bettini et al.} 
%
\tocauthor{Matteo Bettini, Ryan Kortvelesy, Jan Blumenkamp and Amanda Prorok}
\institute{Department of
Computer Science and Technology, University of Cambridge, Cambridge, UK, \email{\{mb2389,rk627,jb2270,asp45\}@cl.cam.ac.uk}}

\maketitle              

\begin{abstract}
While many multi-robot coordination problems can be solved optimally by exact algorithms, solutions are often not scalable in the number of robots. Multi-Agent Reinforcement Learning (MARL) is gaining increasing attention in the robotics community as a promising solution to tackle such problems. Nevertheless, we still lack the tools that allow us to \textit{quickly} and \textit{efficiently} find solutions to large-scale collective learning tasks. In this work, we introduce the Vectorized Multi-Agent Simulator (VMAS). VMAS is an open-source framework designed for efficient MARL benchmarking. It is comprised of a vectorized 2D physics engine written in PyTorch and a set of twelve challenging multi-robot scenarios. Additional scenarios can be implemented through a simple and modular interface. We demonstrate how vectorization enables parallel simulation on accelerated hardware without added complexity. When comparing VMAS to OpenAI MPE, we show how MPE's execution time increases linearly in the number of simulations while VMAS is able to execute 30,000 parallel simulations in under 10s, proving more than 100$\times$ faster. Using VMAS's RLlib interface, we benchmark our multi-robot scenarios using various Proximal Policy Optimization (PPO)-based MARL algorithms. VMAS's scenarios prove challenging in orthogonal ways for state-of-the-art MARL algorithms. The VMAS framework is available at: \url{https://github.com/proroklab/VectorizedMultiAgentSimulator}. A video of VMAS scenarios and experiments is available \href{https://youtu.be/aaDRYfiesAY}{here}\footnote{\url{https://youtu.be/aaDRYfiesAY} \label{foot:video}}
\end{abstract}

\begin{figure}[ht]
    \newcommand{\subfigsize}{0.25}
     \centering
       \begin{subfigure}{\subfigsize\linewidth}
         \centering
         \includegraphics[width=\linewidth,frame]{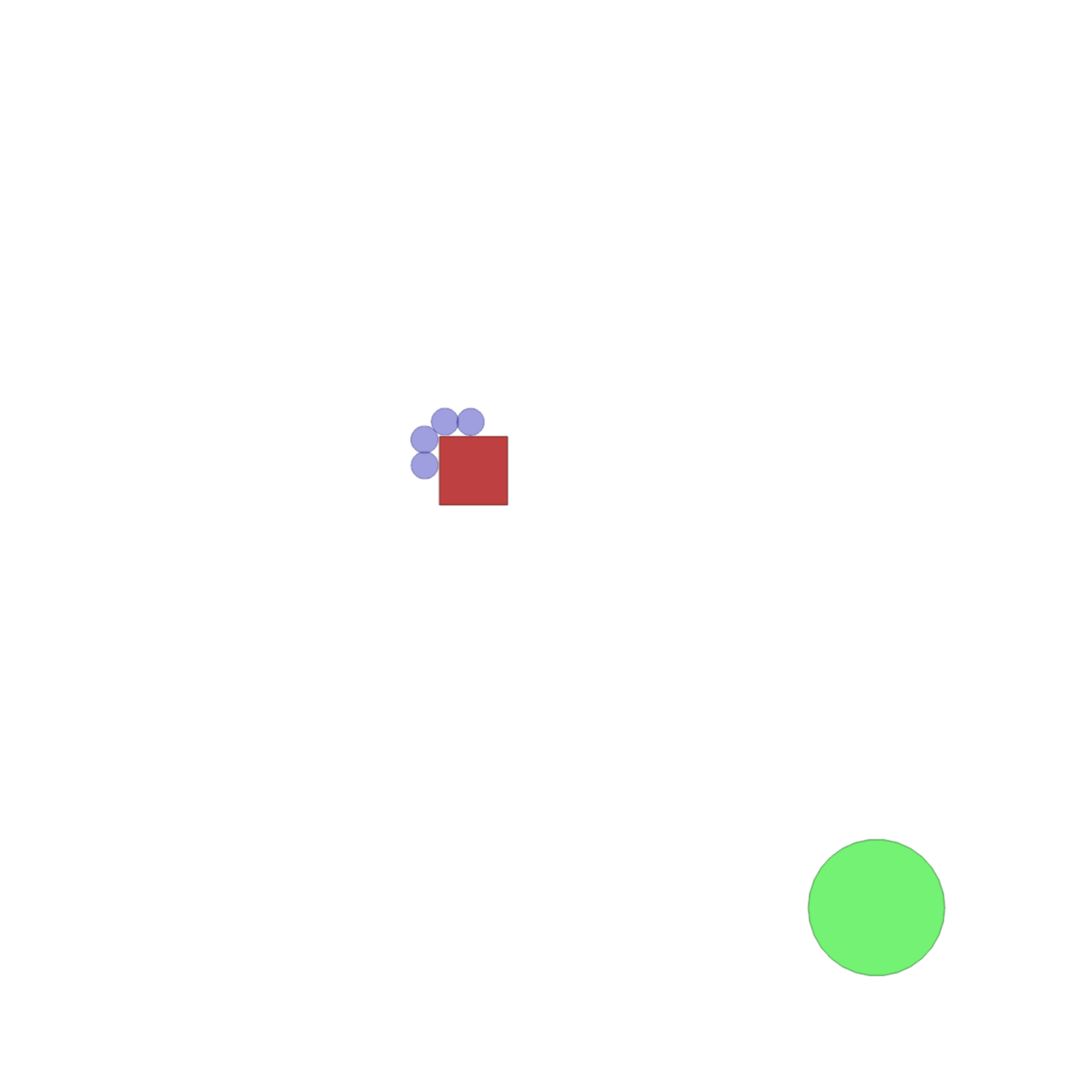}
         \caption{Transport}
         \label{fig:transport}
     \end{subfigure}%
          \begin{subfigure}{\subfigsize\linewidth}
         \centering
         \includegraphics[width=\linewidth,frame]{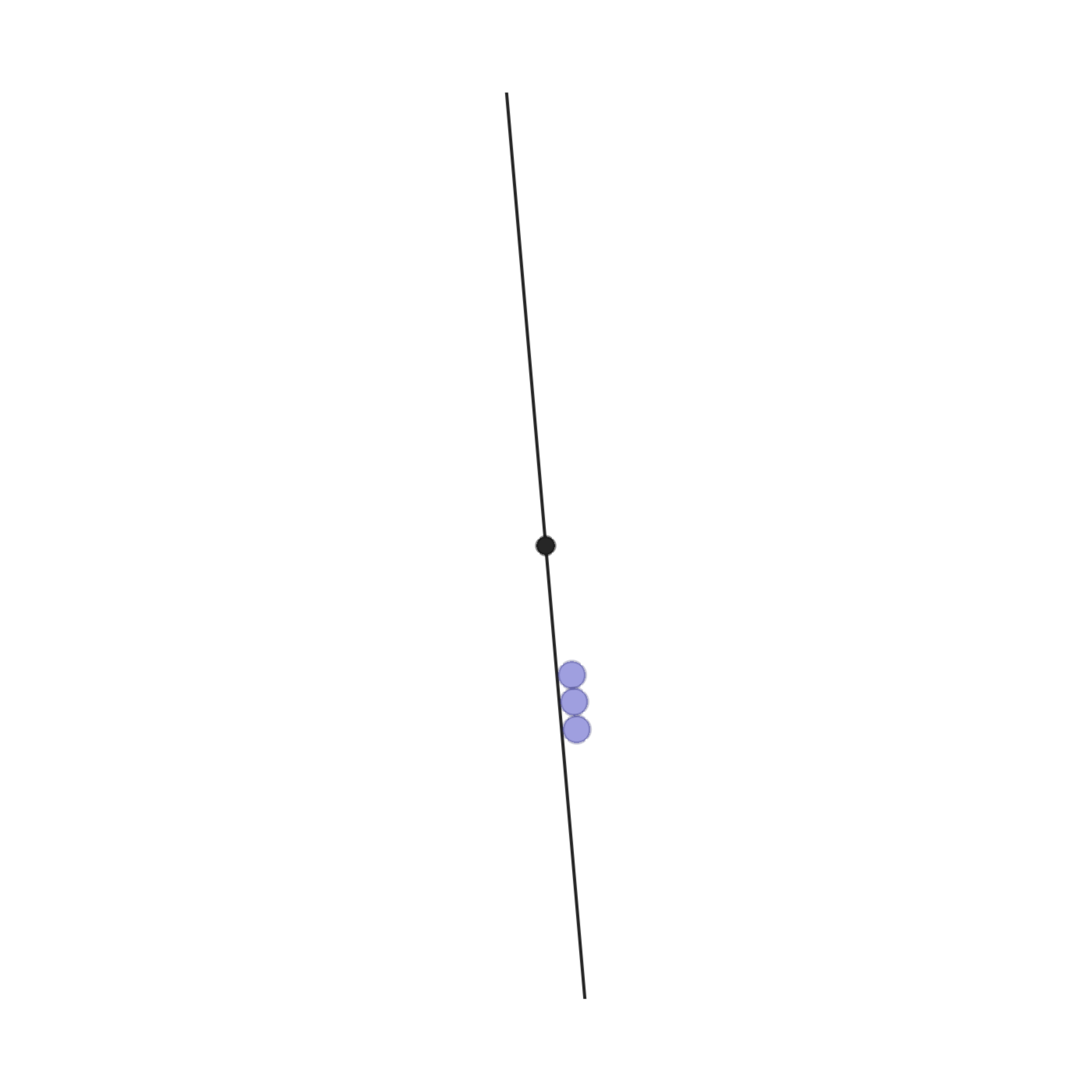}
         \caption{Wheel}
         \label{fig:wheel}
     \end{subfigure}%
     \begin{subfigure}{\subfigsize\linewidth}
         \centering
         \includegraphics[width=\linewidth,frame]{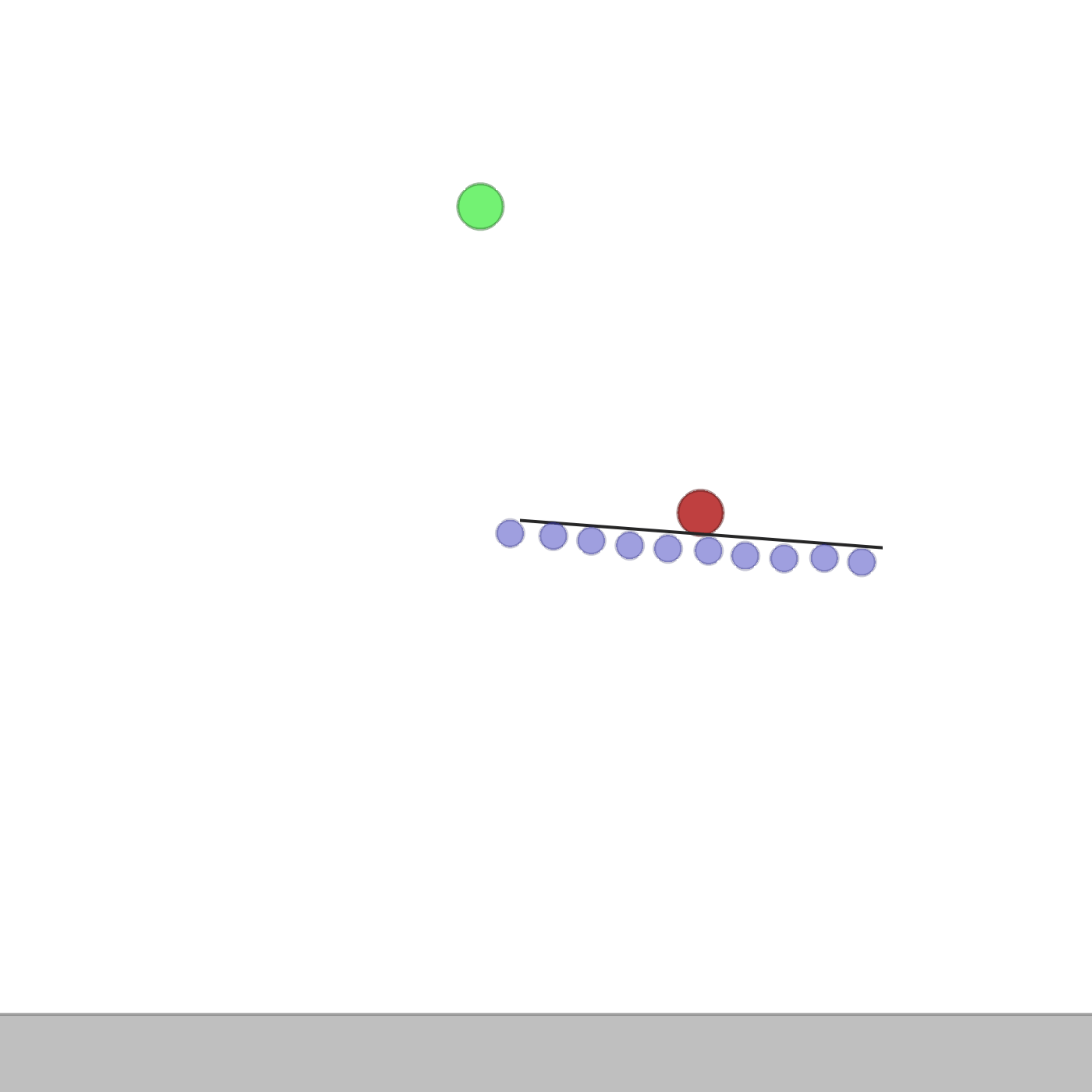}
         \caption{Balance}
         \label{fig:balance}
     \end{subfigure}%
       \begin{subfigure}{\subfigsize\linewidth}
         \centering
         \includegraphics[width=\linewidth,frame]{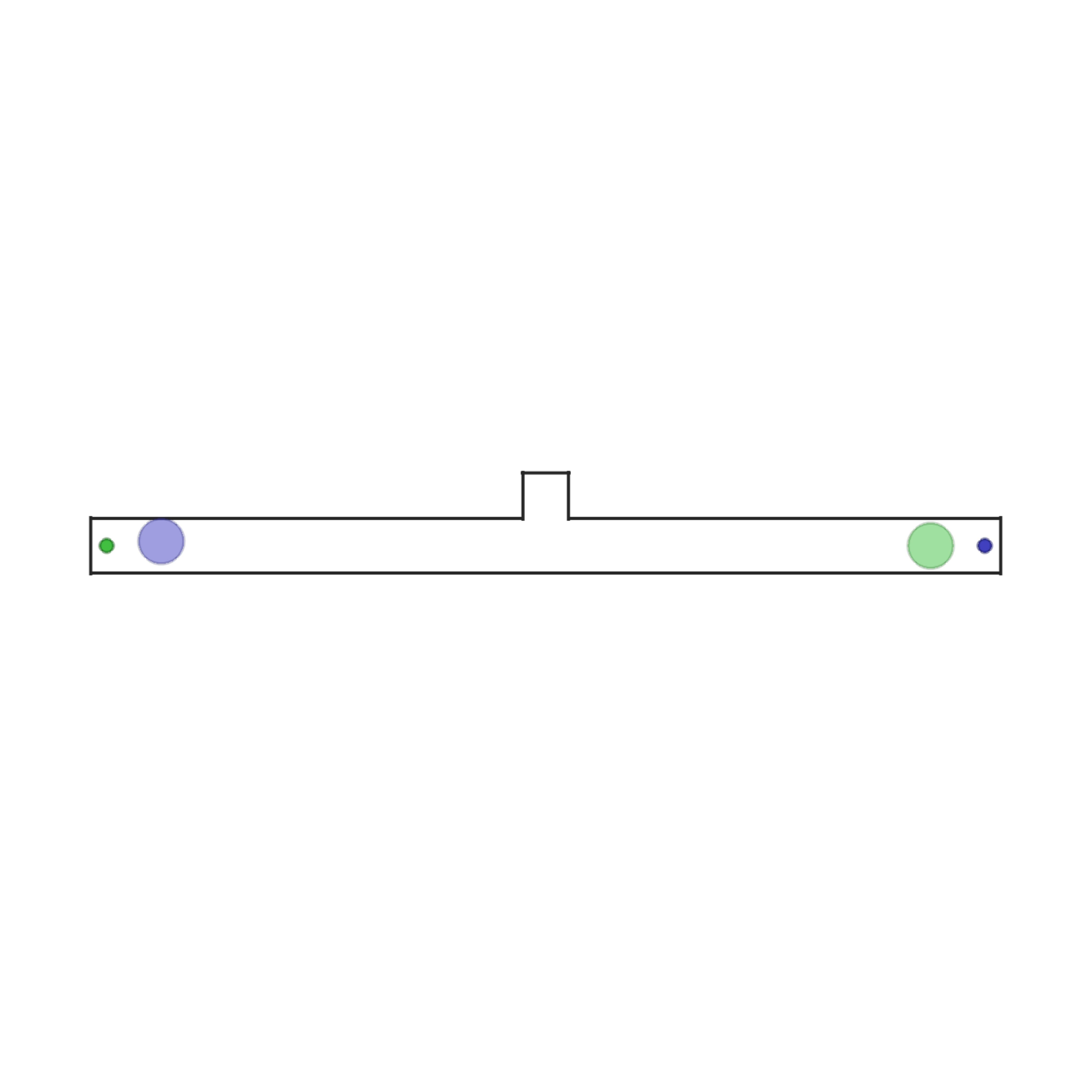}
         \caption{Give way}
         \label{fig:give_way}
     \end{subfigure}
    \begin{subfigure}{\subfigsize\linewidth}
         \centering
         \includegraphics[width=\linewidth,frame]{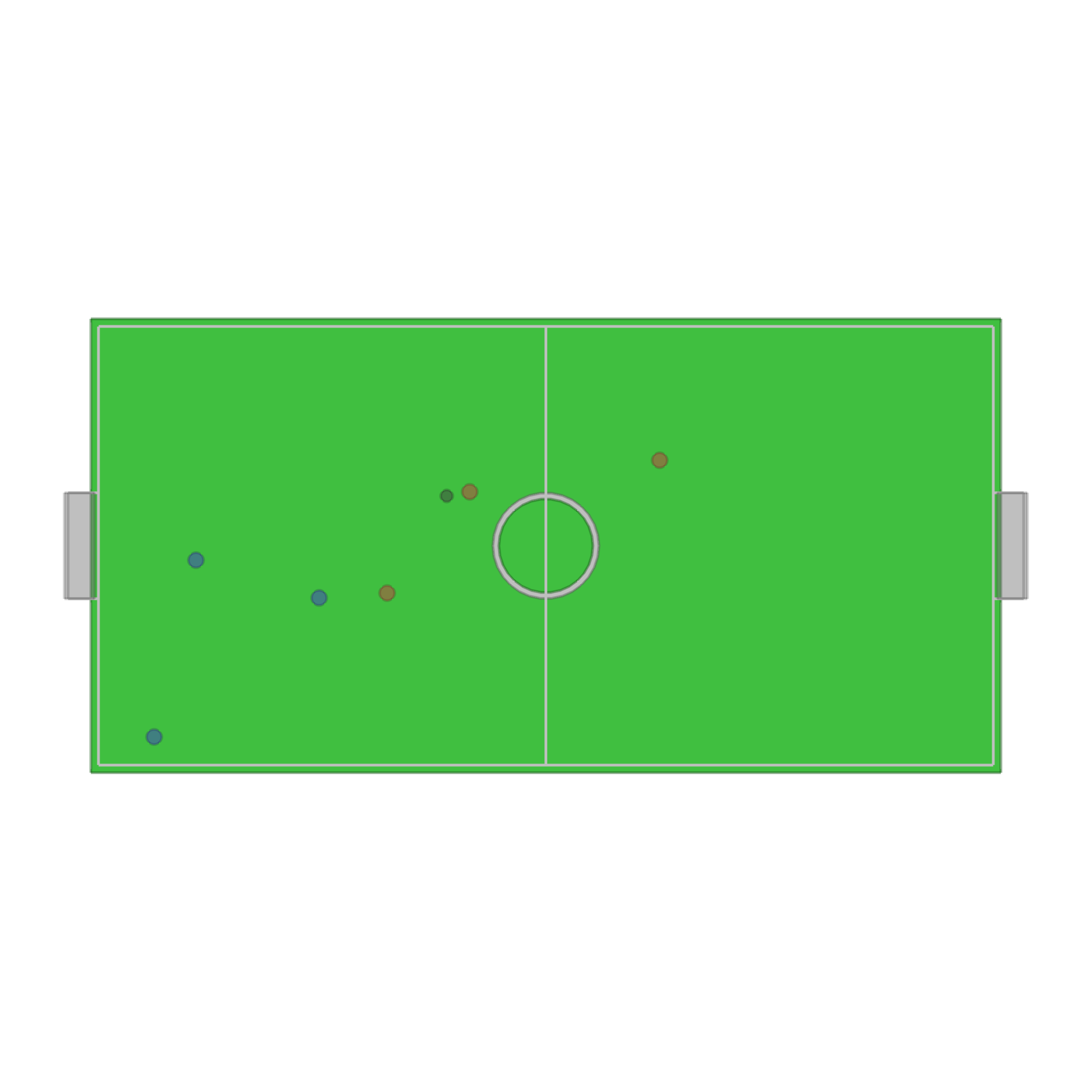}
         \caption{Football}
         \label{fig:football}
     \end{subfigure}%
     \begin{subfigure}{\subfigsize\linewidth}
         \centering
         \includegraphics[width=\linewidth,frame]{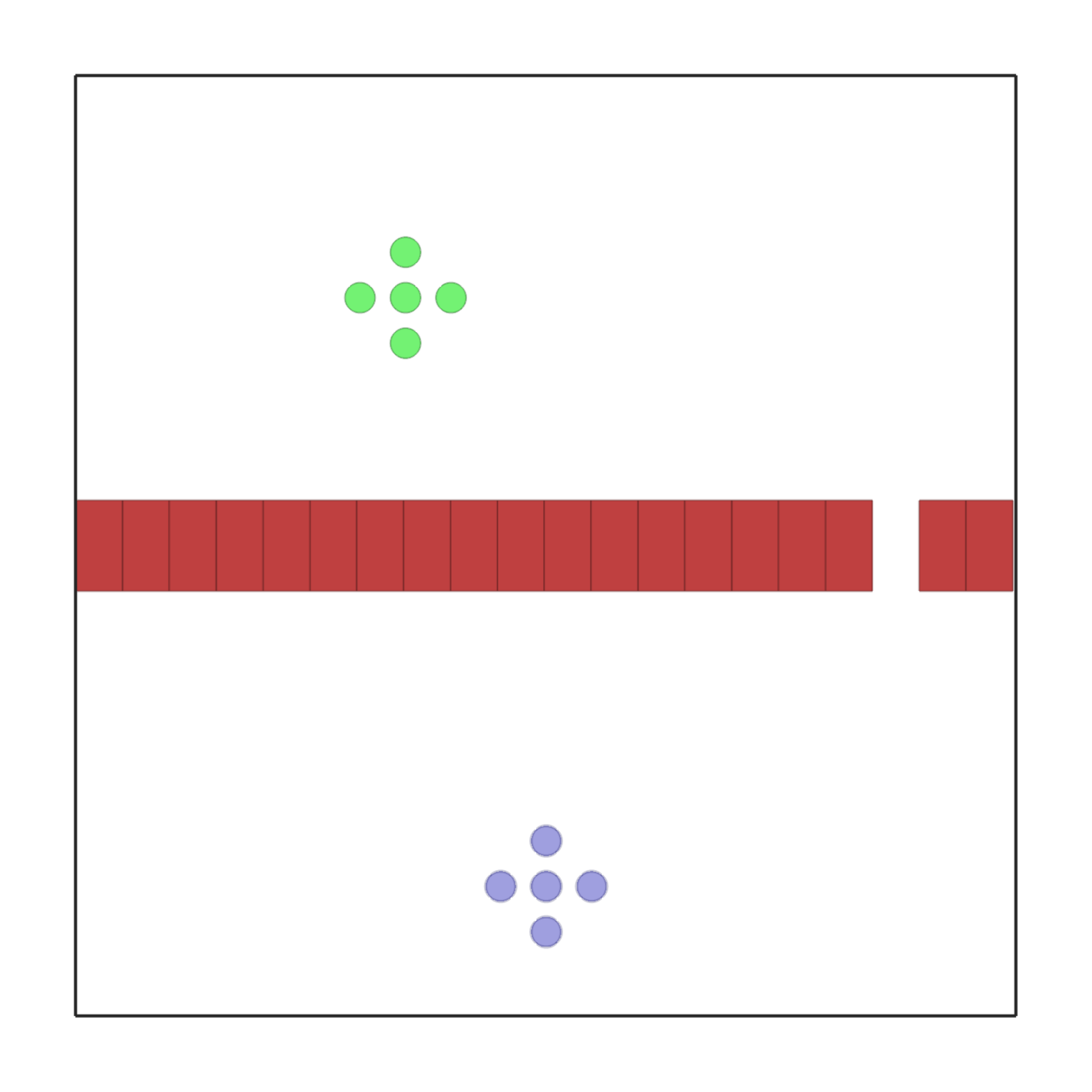}
         \caption{Passage}
         \label{fig:passage}
     \end{subfigure}%
     \begin{subfigure}{\subfigsize\linewidth}
         \centering
         \includegraphics[width=\linewidth,frame]{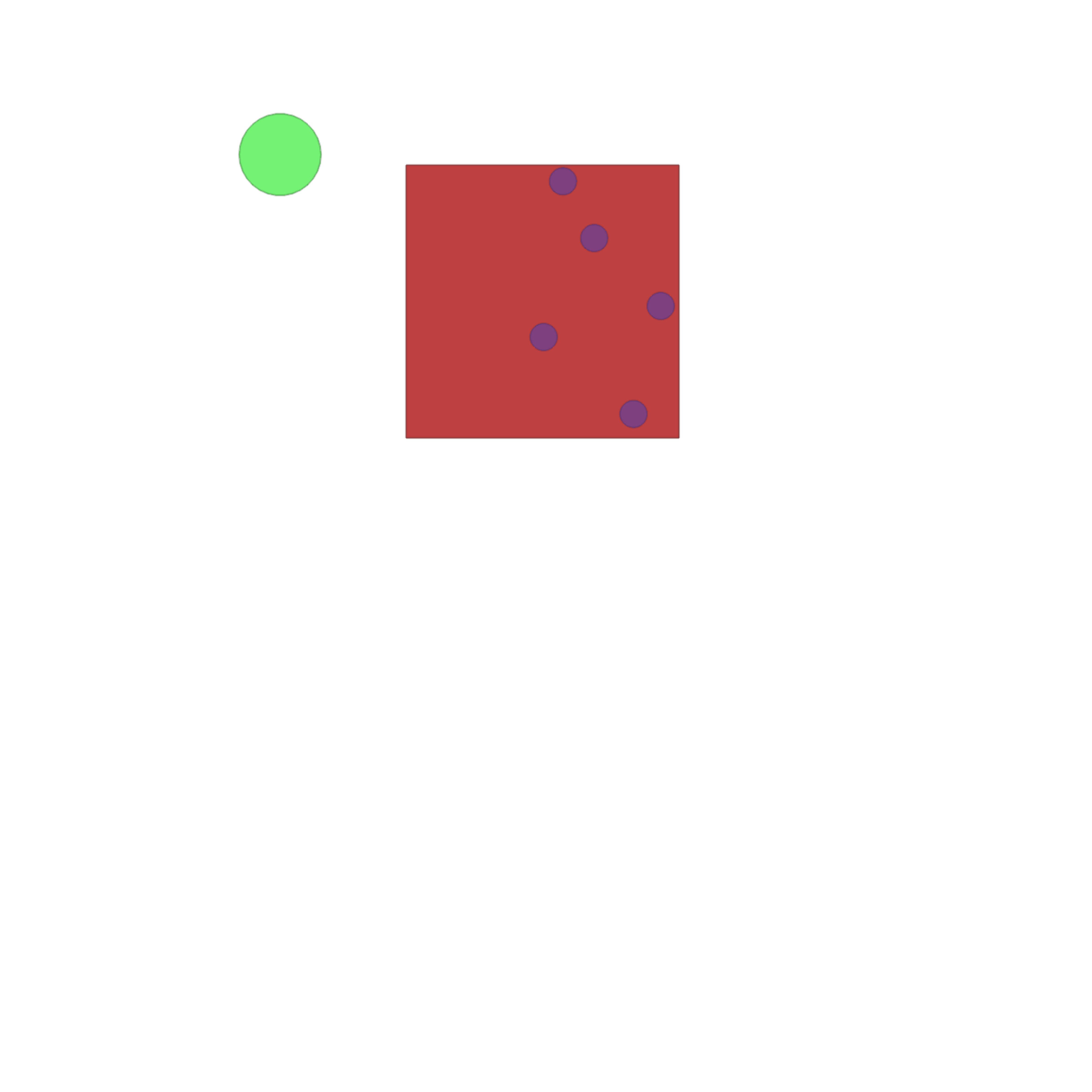}
         \caption{Reverse Transport}
         \label{fig:reverse_transport}
     \end{subfigure}%
     \begin{subfigure}{\subfigsize\linewidth}
         \centering
         \includegraphics[width=\linewidth,frame]{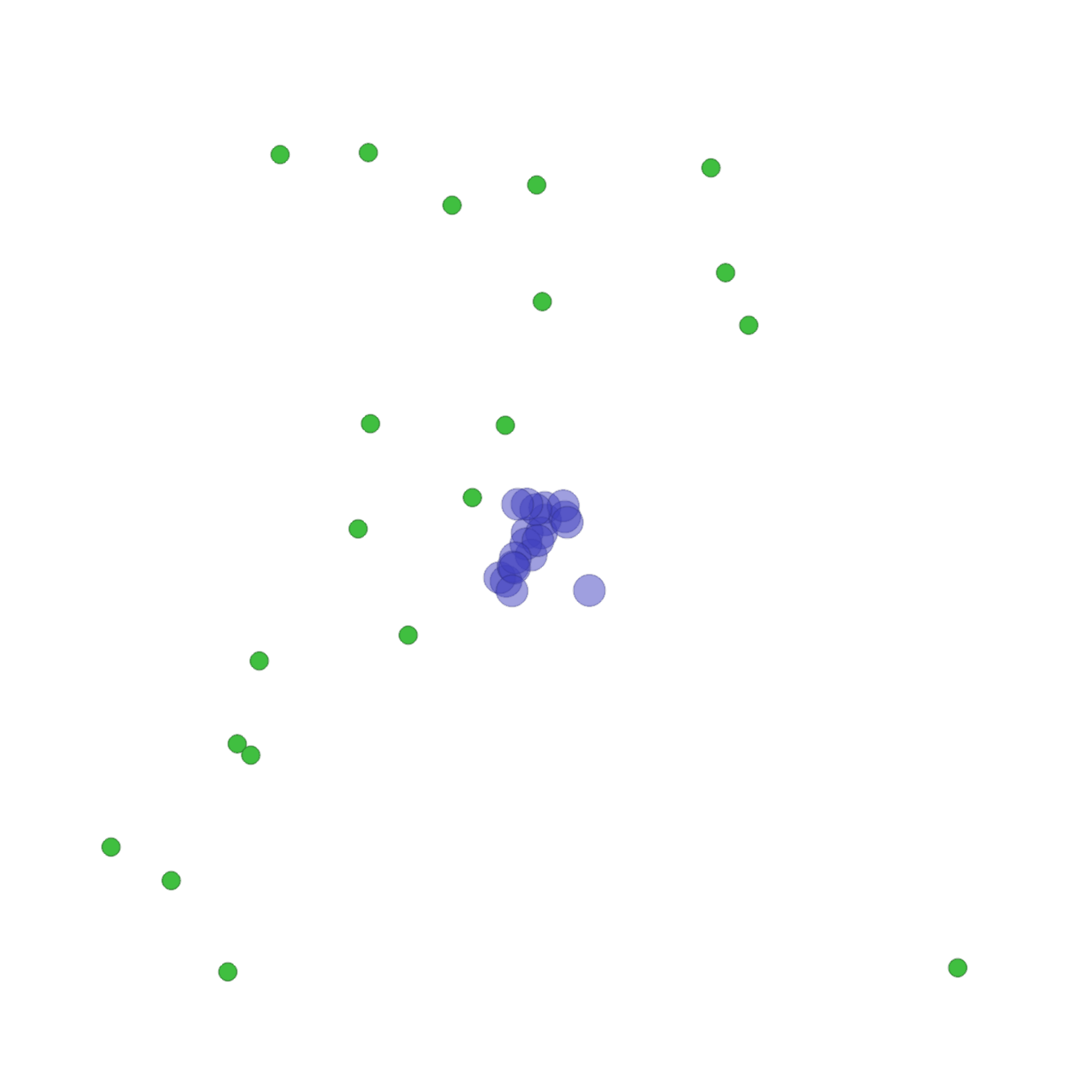}
         \caption{Dispersion}
         \label{fig:dispersion}
     \end{subfigure}
    
     \begin{subfigure}{\subfigsize\linewidth}
         \centering
         \includegraphics[width=\linewidth,frame]{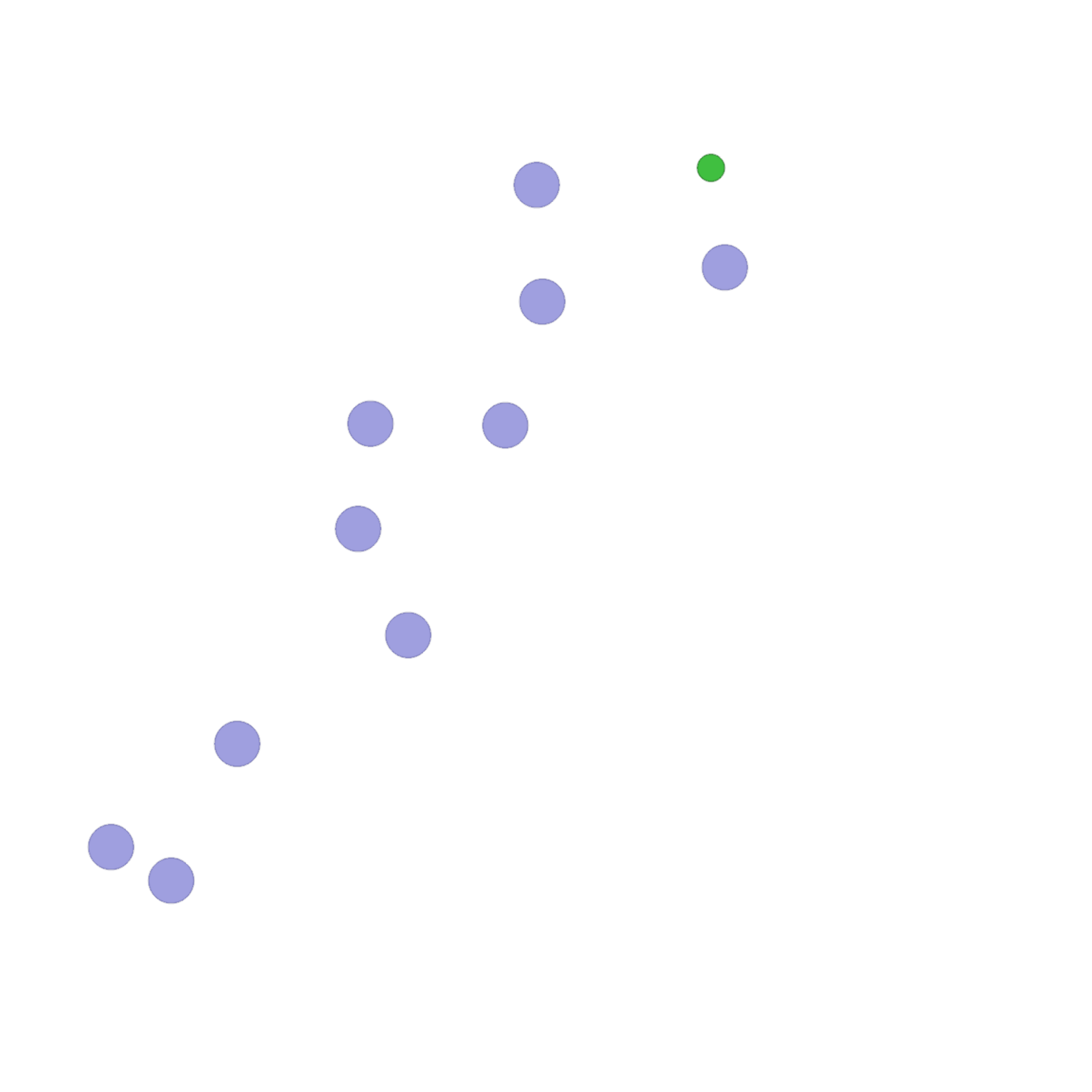}
         \caption{Dropout}
         \label{fig:dropout}
     \end{subfigure}%
     \begin{subfigure}{\subfigsize\linewidth}
         \centering
         \includegraphics[width=\linewidth,frame]{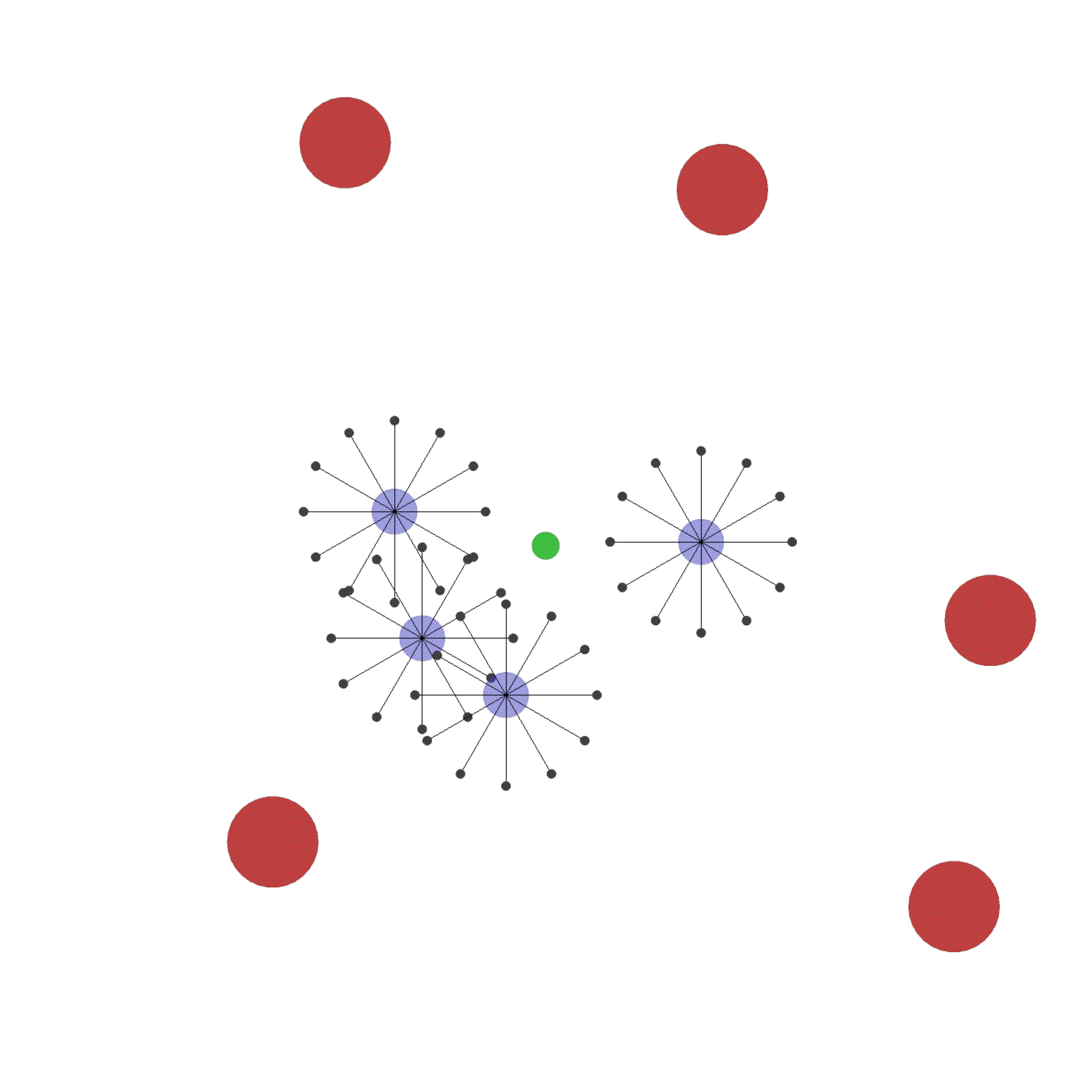}
         \caption{Flocking}
         \label{fig:flocking}
     \end{subfigure}%
     \begin{subfigure}{\subfigsize\linewidth}
         \centering
         \includegraphics[width=\linewidth,frame]{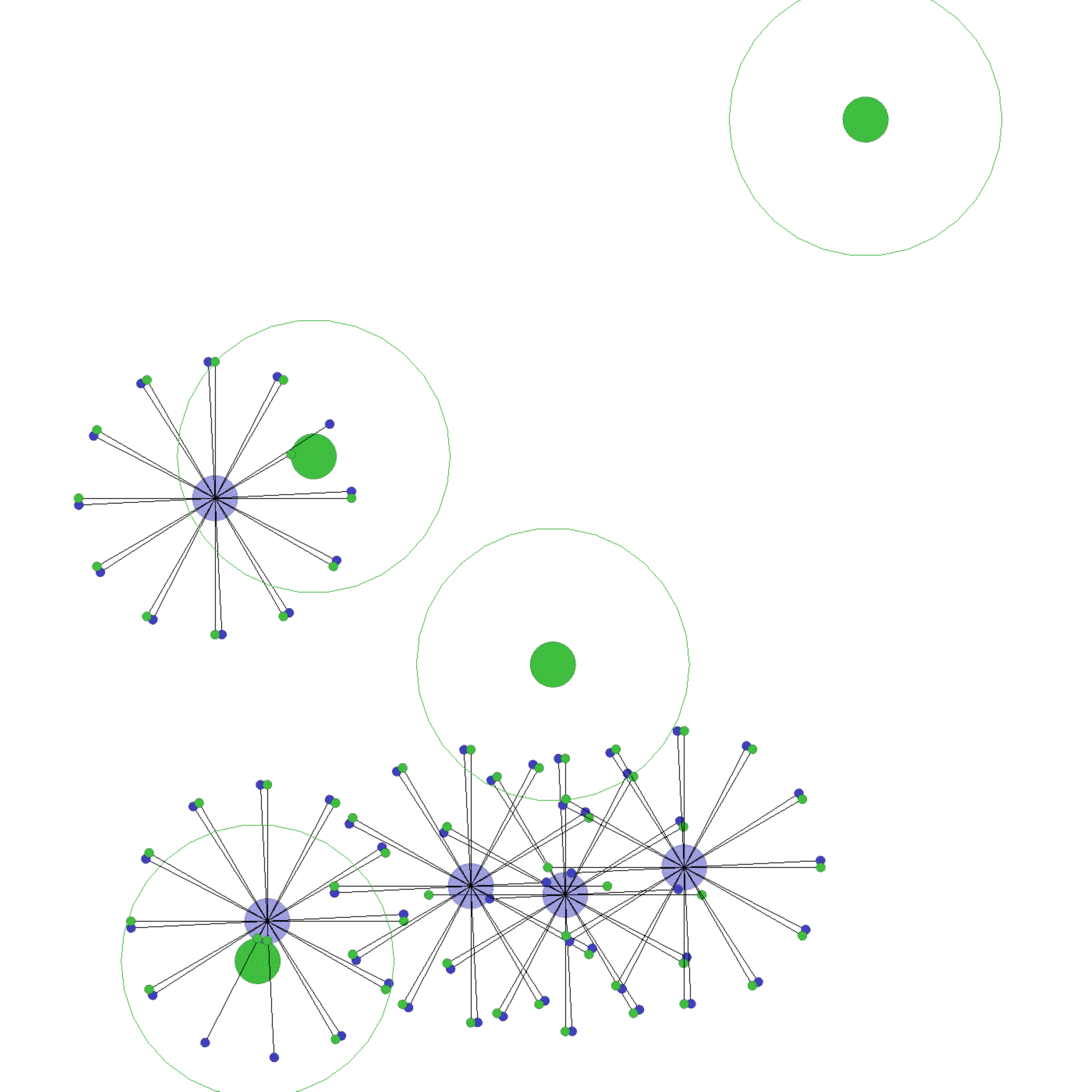}
         \caption{Discovery}
         \label{fig:discovery}
     \end{subfigure}%
     \begin{subfigure}{\subfigsize\linewidth}
         \centering
         \includegraphics[width=\linewidth,frame]{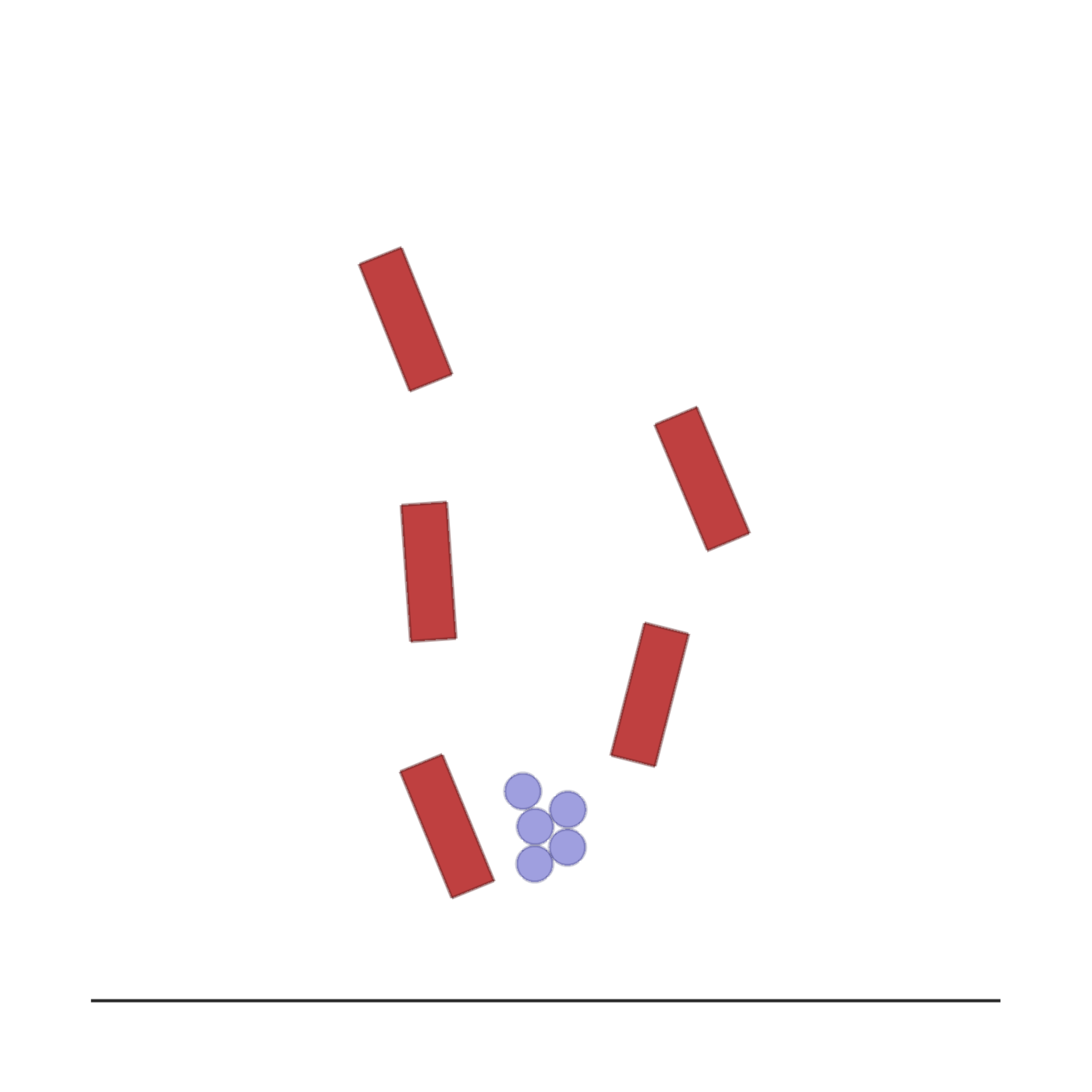}
         \caption{Waterfall}
         \label{fig:waterfall}
     \end{subfigure}%
    \caption{Multi-robot scenarios introduced in VMAS. Robots (blue shapes) interact among each other and with landmarks (green, red, and black shapes) to solve a task.}
    \label{fig:vmaps_scenarios}
\end{figure}

\section{Introduction}
\label{sec:introduction}
\input{sections/introduction}
\section{Related work}
\label{sec:related_work}
\input{sections/related_works}
\section{The VMAS platform}
\label{sec:simulator}
\input{sections/simulator}

\section{Multi-robot scenarios}
\label{sec:scenarios}
\input{sections/scenarios}
\section{Comparison with MPE}
\label{sec:mpe_comparison}
\input{sections/mpe_comparison}
\section{Experiments and benchmarks}
\label{sec:experiments}
\input{sections/experiments}
\section{Conclusion}
\label{sec:conclusion}
\input{sections/conclusion}
\section*{Acknowledgements}
This work was supported by ARL DCIST CRA W911NF-17-2-0181 and European Research Council (ERC) Project 949940 (gAIa). R. Kortvelesy was supported by Nokia Bell Labs through their donation for the Centre of Mobile, Wearable Systems and Augmented Intelligence to the University of Cambridge. J. Blumenkamp acknowledges the support of the ‘Studienstiftung des deutschen Volkes’ and an EPSRC tuition fee grant.

\bibliographystyle{spmpsci} 
\bibliography{bibliography}

\end{document}

%% file: sections/introduction.tex

Many real-world problems require coordination of multiple robots to be solved. However, coordination problems are commonly computationally hard. Examples include path-planning~\cite{li2020graph}, task assignment~\cite{prorok2020robust}, and area coverage~\cite{zheng2010multirobot}. While exact solutions exist, their complexity grows exponentially in the number of robots~\cite{bernstein2002complexity}. Metaheuristics~\cite{braysy2005vehicle} provide a fast and scalable solutions, but lack in optimality. Multi-Agent Reinforcement Learning (MARL) can be used as a scalable approach to find near-optimal solutions to these problems~\cite{wang2020mobile}. In MARL, agents trained in simulation collect experiences by interacting with the environment, and train their policies (typically represented with deep neural networks) through a reward signal.

However, current MARL approaches present several issues. Firstly, the training phase can require significant time to converge to optimal behavior. This is partially due to the sample efficiency of the algorithm, and partially to the computational complexity of the simulator. Secondly, current benchmarks are specific to a predefined task and mostly tackle unrealistic videogame-like scenarios~\cite{samvelyan19smac,suarez2019neural}, far from real-world multi-robot problems. This makes research in this area fragmented, with a new simulation framework being implemented for each new task introduced. Multi-robot simulators, on the other hand, prove to be more general, but their high fidelity and full-stack simulation results in slow performance, preventing their applicability to MARL. Full-stack learning can significantly hinder training performance. Learning can be made more sample-efficient if simulation is used to solve high-level multi-robot coordination problems, while leaving low-level robotic control to first-principles-based methods.

Motivated by these reasons, we introduce VMAS, a vectorized multi-agent simulator. VMAS is a vectorized 2D physics simulator written in PyTorch~\cite{paszke2019pytorch}, designed for efficient MARL benchmarking. It simulates agents and landmarks of different shapes and supports torque, elastic collisions and custom gravity. Holonomic motion models are used for the agents to simplify simulation. Vectorization in PyTorch allows VMAS to perform simulations in a batch, seamlessly scaling to tens of thousands of parallel environments on accelerated hardware. With the term \textit{GPU vectorization} we refer to the Single Instruction Multiple Data (SIMD) execution paradigm available inside a GPU warp. This paradigm permits to execute the same instruction on a set of parallel simulations in a batch. VMAS has an interface compatible with OpenAI Gym~\cite{brockman2016openai} and with the RLlib library~\cite{liang2018rllib}, enabling out-of-the-box integration with a wide range of RL algorithms. VMAS also provides a framework to easily implement custom multi-robot scenarios. Using this framework, we introduce a set of 12 multi-robot scenarios representing difficult learning problems. Additional scenarios can be implemented through a simple and modular interface. We vectorize and port all scenarios from OpenAI MPE~\cite{lowe2017multi} in VMAS. We benchmark four of VMAS's new scenarios using three MARL algorithms based on Proximal Policy Optimization (PPO)~\cite{schulman2017proximal}. We show the benefits of vectorization by benchmarking our scenarios in the RLlib~\cite{liang2018rllib} library. Our scenarios prove to challenge state-of-the-art MARL algorithms in complementary ways.

\textbf{Contributions}. We now list the main contributions of this work:
\begin{itemize}
    \item We introduce the \textit{VMAS framework}. A vectorized multi-agent simulator which enables MARL training at scale. VMAS supports inter-agent communication and customizable sensors, such as LIDARs.
    \item We implement a set of \textit{twelve multi-robot scenarios} in VMAS, which focus on testing different collective learning challenges including: behavioural heterogeneity, coordination through communication, and adversarial interaction. 
    \item We port and vectorize all scenarios from OpenAI MPE~\cite{lowe2017multi} into VMAS and run a performance comparison between the two simulators. We demonstrate the benefits of vectorization in terms of simulation speed, showing that VMAS is up to 100$\times$ faster than MPE.
\end{itemize}
\noindent The VMAS codebase is available \href{https://github.com/proroklab/VectorizedMultiAgentParticleSimulator}{here}\footnote{\url{https://github.com/proroklab/VectorizedMultiAgentSimulator}\label{foot:vmas_url}}.

%% file: sections/related_works.tex
In this section, we review the related literature in the fields of multi-agent and multi-robot simulation, highlighting the core gaps of each field. Furthermore, we compare the most relevant simulation frameworks with VMAS in \autoref{tab:simulator_comparison}. 

\textbf{Multi-agent reinforcement learning environments}. A significant amount of work exists in the context of MARL to address the issues of multi-robot simulation for learning hard coordination strategies. Realistic GPU-accelerated simulators and engines have been proposed. Isaac~\cite{makoviychuk2021isaac} is a proprietary NVIDIA simulator used for realistic robotic simulation in reinforcement learning. Instead of using environment vectorization to accelerate learning, it uses concurrent execution of multiple training environments in the same simulation instance. Despite of this, its high-fidelity simulation makes it computationally expensive for high-level MARL problems. Brax~\cite{brax2021github} is a vectorized 3D physics engine introduced by Google. It uses the Jax~\cite{jax2018github} library to achieve environment batching and full-differentiability. However, computational issues occur when scaling the number of simulated agents, leading to stalled environments with just 20 agents. There also exist projects for single-agent vectorized environments~\cite{gymnax2022github,envpool}, but the complexity of extending these to the multi-agent domain is non-trivial.

The core benchmark environments of the MARL literature focus on high-level inter-robot learning. Multiagent Particle Environments (MPE)~\cite{lowe2017multi} are a set of enviroments created by OpenAI. They share VMAS's principles of modularity and ease of new scenario creation, without providing environment vectorization. \text{MAgent}~\cite{zheng2018magent} is a discrete-world environment supporting a high number of agents. Multi-Agent-Learning-Environments~\cite{jiang2021multi} is another simplified discrete-world set of environments with a range of different multi-robot tasks. Multi-Agent-Emergence-Environments~\cite{baker2019emergent} is a customizable OpenAI 3D simulator for hide-and-seek style games. Pommerman~\cite{DBLP:journals/corr/abs-1809-07124} is a discretized playground for learning multi-agent competitive strategies.
SMAC~\cite{samvelyan19smac} is a very popular MARL benchmark based on the Starcraft 2 videogame. Neural-MMO~\cite{suarez2019neural} is another videogame-like set of environments where agents learn to survive in large populations. Google Research Football~\cite{kurach2020google} is a football simulation with a suite of scenarios that test different aspects of the game.
Gym-pybullet-drones~\cite{panerati2021learning} is a realistic PyBullet simulator for multi-quadricopters control. Particle Robots Simulator~\cite{shen2022deep} is a simulator for particle robots, which require high coordination strategies to overcome actuation limitations and achieve high-level tasks. Multi-Agent Mujoco~\cite{peng2021facmac} consists in multiple agents controlling different body parts of a single Mujoco~\cite{todorov2012mujoco} agent. While all these environments provide interesting MARL benchmarks, most of them focus on specific tasks. Furthermore, none of these environments provide GPU vectorization, which is key for efficient MARL training. We present a comparison between VMAS and all the aforementioned environments in \autoref{tab:simulator_comparison}.

\textbf{Multi-robot simulators}. Video-game physics engines such as Unity and Unreal Engine grant realistic simulation that can be leveraged for multi-agent robotics. Both make use of the GPU-accelerated NVIDIA PhysX. However, their generality causes high overheads when using them for robotics research.
Other popular physics engines are Bullet, Chipmunk, Box2D, and ODE. These engines are all similar in their capabilities and prove easier to adopt due to the availability of Python APIs. Thus, they are often the tool of choice for realistic robotic simulation. However, because they do not leverage GPU-accelerated batched simulation, these tools lead to performance bottlenecks in MARL training.

The most widely known robotic simulators are Gazebo~\cite{koenig2004design} and Webots~\cite{michel2004cyberbotics}. Their engines are based on the ODE 3D dynamics library. These simulators support a wide range of robot models, sensors, and actuators, but suffer from significant performance loss when scaling in the number of agents. Complete simulation stall is shown to occur with as few as 12 robots~\cite{8088134}. For this reason, Argos~\cite{Pinciroli:SI2012} has been proposed as a scalable multi-robot simulator. It is able to simulate swarms in the thousands of agents by assigning parts of the simulation space to different physics engines with different simulation goals and fidelity. Furthermore, it uses CPU parallelization through multi-threading. Despite these features, none of the simulators described are fast enough to be usable in MARL training. This is because they prioritize realistic full-stack multi-robot simulation over speed, and they do not leverage GPU acceleration for parallel simulations. 
This focus on realism is not always necessary in MARL. In fact, most collective coordination problems can be decoupled from low-level problems relating to sensing and control. When these problems can be efficiently solved independently without loss of generality, fast high-level simulation provides an important tool. This insight is the key factor motivating the holonomicity assumption in VMAS.

\begin{table}[t]
\caption{Comparison of multi-agent and multi-robot simulators and environments.}
\label{tab:simulator_comparison} 
\resizebox{\linewidth}{!}{%
\begin{tabular}{r c c c c c c c c c c c}
\hline\noalign{\smallskip}
 & \rotatebox{0}{Vector$^a$} & \rotatebox{0}{State$^b$} & \rotatebox{0}{Comm$^c$} & \rotatebox{0}{Action$^d$} & \rotatebox{0}{PhysEng$^e$} & \rotatebox{0}{\#Agents$^f$} & \rotatebox{0}{Gen$^g$} & \rotatebox{0}{Ext$^h$} & \rotatebox{0}{MRob$^i$} & \rotatebox{0}{MARL$^j$} & \rotatebox{0}{RLlib$^k$} \\
\toprule

Brax~\cite{brax2021github} & \cmark  & C & \xmark & C & 3D & $<10$ & \cmark & \cmark & \xmark & \xmark &\xmark\\
MPE~\cite{lowe2017multi}  & \xmark  & C & C+D & C+D & 2D & $<100$ & \cmark & \cmark & \xmark & \cmark &\cmark\\
MAgent~\cite{zheng2018magent} & \xmark  & D & \xmark & D & \xmark & $>1000$ & \xmark & \xmark & \xmark & \cmark &\cmark\\
MA-Learning-Environments~\cite{jiang2021multi} & \xmark  & D & \xmark & D & \xmark & $<10$ & \cmark & \xmark & \cmark & \cmark &\xmark\\
MA-Emergence-Environments~\cite{baker2019emergent} & \xmark  & C & \xmark & C+D & 3D & $<10$ & \xmark & \xmark & \xmark & \cmark &\xmark\\
Pommerman~\cite{DBLP:journals/corr/abs-1809-07124} & \xmark  & D & \xmark & D & \xmark & $<10$ & \xmark & \xmark & \xmark & \cmark &\xmark\\
SMAC~\cite{samvelyan19smac}  & \xmark  & C & \xmark & D & \xmark & $<100$ & \xmark & \cmark & \xmark & \cmark & \cmark\\
Neural-MMO~\cite{suarez2019neural} & \xmark  & C & \xmark & C+D & \xmark & $<1000$ & \xmark & \cmark & \xmark & \cmark & \cmark \\
Google research football~\cite{kurach2020google} & \xmark  & C & \xmark & D & 2D & $<100$ & \xmark & \cmark & \xmark & \cmark &\cmark\\
gym-pybullet-drones~\cite{panerati2021learning} & \xmark & C & \xmark & C & 3D & $<100$ & \xmark & \cmark &\cmark & \cmark &\cmark \\
Particle robots simulator~\cite{shen2022deep} & \xmark  & C & \xmark & C+D & 2D & $<100$ & \xmark & \cmark & \cmark & \cmark &\xmark\\
MAMujoco~\cite{peng2021facmac} & \xmark  & C & \xmark & C & 3D & $<10$ & \xmark & \xmark & \xmark & \cmark &\xmark\\

\midrule

Gazebo~\cite{koenig2004design} &  \xmark & C & C+D & C+D & 3D & $<10$  & \cmark & \cmark & \cmark & \xmark & \xmark\\
Webots~\cite{michel2004cyberbotics} & \xmark  & C & C+D  & C+D & 3D & $<10$  & \cmark & \cmark & \cmark &\xmark  & \xmark \\
ARGOS~\cite{Pinciroli:SI2012} & \xmark & C & C+D & C+D & 2D\&3D & $<1000$ & \cmark & \cmark &\cmark & \xmark &\xmark \\

\midrule

VMAS & \cmark  & C & C+D & C+D & 2D & $<100$ & \cmark & \cmark & \cmark & \cmark &\cmark\\

\bottomrule
\end{tabular}}\\

$^a$ Vectorized\\
$^b$ Continuous state (C) or discrete state/grid world (D)\\
$^c$ Continuous communication (C) or discrete communication (D) inside the simulator\\
$^d$ Continuous actions (C) or discrete actions (D)\\
$^e$ Type of physics engine\\
$^f$ Number of agents supported\\
$^g$ General purpose simulator: any type of task can be created\\
$^h$ Extensibility (API for creating new scenarios)\\
$^i$ Contains multi-robot tasks\\
$^j$ Made for Multi-Agent Reinforcement Learning (MARL)\\
$^k$ Compatible with RLlib framework~\cite{liang2018rllib}
\end{table}

%% file: sections/simulator.tex
The unique characteristic that makes VMAS different from the related works compared in \autoref{tab:simulator_comparison} is the fact that our platform brings together multi-agent learning and environment vectorization. Vectorization is a key component to speed-up MARL training. In fact, an on-policy training iteration\footnote{Here we illustrate an on-policy training iteration, but simulation is a key component of any type of MARL algorithm} is comprised of simulated team rollouts and a policy update. During the rollout phase of iteration $k$, simulations are performed to collect experiences from the agents' interactions with the environment according to their policy $\pi_k$. The collected experiences are then used to update the team policy. The new policy $\pi_{k+1}$ will be employed in the rollout phase of the next training iteration. The rollout phase usually constitutes the bottleneck of this process. Vectorization allows parallel simulation and helps alleviate this issue.

Inspired by the modularity of some existing solutions, like MPE~\cite{lowe2017multi}, we created our framework as a new scalable platform for running and creating MARL benchmarks. With this goal in mind, we developed VMAS following a set of tenets:

\begin{itemize}
    \item \textbf{Vectorized}. VMAS vectorization can step any number of environments in parallel. This significantly reduces the time needed to collect rollouts for training in MARL.
    \item \textbf{Simple}. Complex vectorized physics engines exist (e.g., Brax~\cite{brax2021github}), but they do not scale efficiently when dealing with multiple agents. This defeats the computational speed goal set by vectorization. VMAS uses a simple custom 2D dynamics engine written in PyTorch to provide fast simulation. 
    \item \textbf{General}. The core of VMAS is structured so that it can be used to implement general high-level multi-robot problems in 2D. It can support adversarial as well as cooperative scenarios. Holonomic robot simulation shifts focus to high-level coordination, obviating the need to learn low-level controls using MARL.
    \item \textbf{Extensible}. VMAS is not just a simulator with a set of environments. It is a framework that can be used to create new multi-agent scenarios in a format that is usable by the whole MARL community. For this purpose, we have modularized our framework to enable new task creation and introduced interactive rendering to debug scenarios.
    \item \textbf{Compatible}. VMAS has multiple wrappers which make it directly compatible with different MARL interfaces, including RLlib~\cite{liang2018rllib} and Gym~\cite{brockman2016openai}. RLlib has a large number of already implemented RL algorithms.

\end{itemize}
Let us break down VMAS's structure in depth.

\begin{figure}[ht]
\centering
\includegraphics[width=0.85\textwidth]{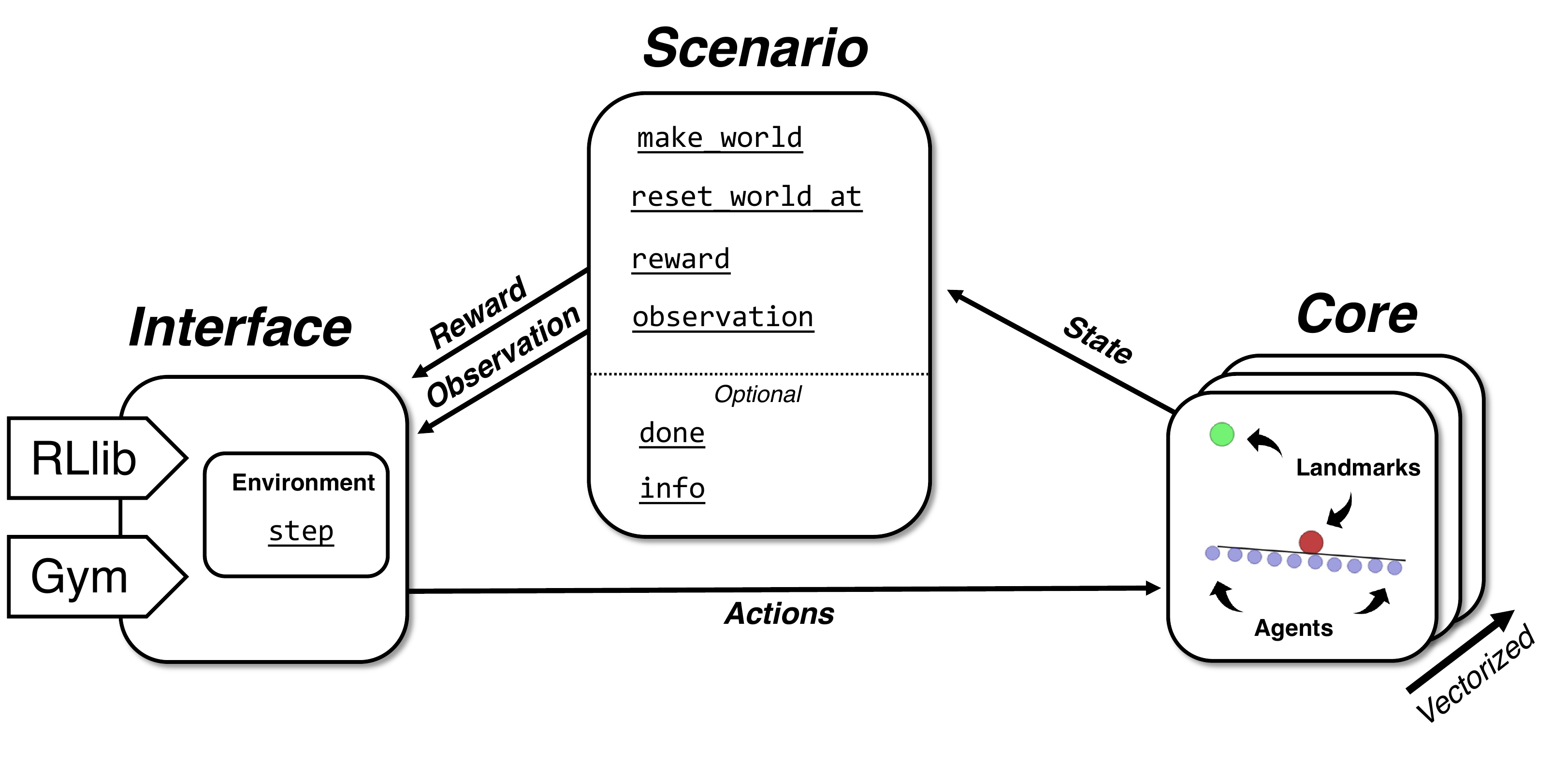}
\caption{VMAS structure. VMAS has a vectorized MARL interface (left) with wrappers for compatibility with OpenAI Gym~\cite{brockman2016openai} and the RLlib RL library~\cite{liang2018rllib}. The default VMAS interface uses PyTorch~\cite{paszke2019pytorch} and can be used for feeding input already on the GPU. Multi-agent tasks in VMAS are defined as scenarios (center). To define a scenario, it is sufficient to implement the listed functions. Scenarios access the VMAS core (right), where agents and landmarks are simulated in the world using a 2D custom written physics module.}
\label{fig:vmaps_structure}
\end{figure}

\textbf{Interface}. The structure of VMAS is illustrated in \autoref{fig:vmaps_structure}. It has a vectorized interface, which means that an arbitrary number of environments can be stepped in parallel in a batch. In \autoref{sec:mpe_comparison}, we demonstrate how vectorization grants important speed-ups on the CPU and seamless scaling on the GPU. While the standard simulator interface uses PyTorch~\cite{paszke2019pytorch} to enable feeding tensors directly as input/output, we provide wrappers for the standard non-vectorized OpenAI Gym~\cite{brockman2016openai} interface and for the vectorized interface of the RLlib~\cite{liang2018rllib} framework. This enables  users to effortlessly access the range of RL training algorithms already available in RLlib. Actions for all environments and agents are fed to VMAS for every simulation step. VMAS supports movement and inter-agent communication actions, both of which can be either continuous or discrete. The interface of VMAS provides rendering through Pyglet~\cite{pyglet}. 

\textbf{Scenario}. Scenarios encode the multi-agent task that the team is trying to solve. Custom scenarios can be implemented in a few hours and debugged using interactive rendering. Interactive rendering is a feature where agents in scenarios can be controlled by users in a videogame-like fashion and all environment-related data is printed on screen. To implement a scenario, it is sufficient to define a few functions: \verb|make_world| creates the agents and landmarks for the scenario and spawns them in the world, \verb|reset_world_at| resets a specific environment in the batch or all environments at the same time, \verb|reward| returns the reward for one agent for all environments, \verb|observation| returns the agent's observations for all environments. Optionally, \verb|done| and \verb|info| can be implemented to provide an ending condition and extra information. Further documentation on how to create new scenarios is available in the \href{https://github.com/proroklab/VectorizedMultiAgentParticleSimulator}{repository}\footref{foot:vmas_url} and in the code.

\textbf{Core}. Scenarios interact with the core. This is where the world simulation is stepped. The world contains $n$ entities, which can be agents or landmarks. Entities have a shape (sphere, box, or line) and a vectorized state $(\mathbf{x}_i,\dot{\mathbf{x}}_i,\theta_i,\dot{\theta}_i ),\, \forall i \in [1..n] \equiv N$, which contains their position $\mathbf{x}_i\in\R^2$, velocity $\dot{\mathbf{x}}_i\in\R^2$, rotation $\theta_i\in\R$, and angular velocity $\dot{\theta}_i \in\R$ for all environments. Entities have a mass $m_i\in\R$ and a maximum speed and can be customized to be movable, rotatable, and collidable. Agents’ actions consist of physical actions, represented as forces $\mathbf{f}^a_i \in \R^2$, and optional communication actions. In the current state of the simulator, agents cannot control their orientation. Agents can either be controlled from the interface or by an “action script” defined in the scenario. Optionally, the simulator can introduce noise to the actions and observations. Custom sensors can be added to agents. We currently support LIDARs.
The world has a simulation step $\delta t$, velocity damping coefficient $\zeta$, and customizable gravity $\mathbf{g} \in \R^2$.

VMAS has a force-based physics engine. Therefore, the simulation step uses the forces at time $t$ to integrate the state by using a semi-implicit Euler method~\cite{niiranen1999fast}:

\begin{equation}
    \begin{cases}
      \mathbf{f}_i(t) = \mathbf{f}^a_i(t) + \mathbf{f}_i^g + \sum_{j \in N \setminus \{i\}}\mathbf{f}_{ij}^e(t) \\
      \dot{\mathbf{x}}_i(t + 1) = (1-\zeta)\dot{\mathbf{x}}_i(t) + \frac{\mathbf{f}_i(t)}{m_i}\delta t\\
      \mathbf{x}_i(t + 1) = \mathbf{x}_i(t) + \dot{\mathbf{x}}_i(t + 1)\delta t 
    \end{cases}\,,
\end{equation}
where $\mathbf{f}^a_i$ is the agent action force, $\mathbf{f}_i^g = m_i\mathbf{g}$ is the force deriving from gravity and $\mathbf{f}_{ij}^e$ is the environmental force used to simulate collisions between entities $i$ and $j$. It has the following form:

\begin{equation}
\mathbf{f}^e_{ij}(t) = 
\begin{cases}
    c \frac{\mathbf{x}_{ij}(t)}{\left \| \mathbf{ x}_{ij}(t)\right \|}  k\log{\left(1 + e^{\frac{-\left(\left \| \mathbf{ x}_{ij}(t)\right \|-d_{\textrm{min}}\right)}{k}}\right )} & \quad\text{if }\left \| \mathbf{ x}_{ij}(t)\right \| \leqslant d_{\textrm{min}} \\
    0  & \quad\text{otherwise}\

    \end{cases}\, .
\end{equation}
 Here, $c$ is a parameter regulating the force intensity. $\mathbf{x}_{ij}$ is the relative position between the closest points on the shapes of the two entities. $d_{\textrm{min}}$ is the minimum distance allowable between them. The term inside the logarithm computes a scalar proportional to the penetration of the two entities, parameterized by a coefficient $k$. This term is then multiplied by the normalized relative position vector. Collision intensity and penetration can be tuned by regulating $c$ and $k$. This is the same collision system used in OpenAI MPE~\cite{lowe2017multi}. 
 
 The simulation step used for the linear state is also applied to the angular state:

\begin{equation}
    \begin{cases}
      \tau_i(t) =  \sum_{j \in N \setminus \{i\}}\left \| \mathbf{r}_{ij}(t) \times \mathbf{f}^e_{ij}(t) \right \| \\
      \dot{\theta}_i(t + 1) = (1-\zeta)\dot{\theta}_i(t) + \frac{\tau_i(t)}{I_i}\delta t\\
      \theta_i(t+1) = \theta_i(t) + \dot{\theta}_i(t+1)\delta t 
    \end{cases}\,.
\end{equation}
Here, $\mathbf{r}_{ij}\in\R^2$ is the vector from the center of the entity to the colliding point, $\tau_i$ is the torque, and $I_i$ is the moment of inertia of the entity. The rules regulating the physics simulation in the  core are basic 2D dynamics implemented in a vectorized manner using PyTorch. They simulate holonomic (unconstrained motion) entities only. 




%% file: sections/scenarios.tex
Alongside VMAS, we introduce a set of 12 multi-robot scenarios. These scenarios contain various multi-robot problems, which require complex coordination---like leveraging heterogeneous behaviour and inter-agent communication---to be solved. While the ability to send communication actions is not used in these scenarios, communication can be used in the policy to improve performance. For example, Graph Neural Networks (GNNs) can be used to overcome partial observability through information sharing~\cite{blumenkamp2021framework}. 

Each scenario delimits the agents' input by defining the set of their observations. This set typically contains the minimum observation needed to solve the task (e.g., position, velocity, sensory input, goal position).
Scenarios can be made arbitrarily harder or easier by modifying these observations. For example, if the agents are trying to transport a package, the precise relative distance to the package can be removed from the agent inputs and replaced with LIDAR measurements. Removing global observations from a scenario is a good incentive for inter-agent communication.

All tasks contain numerous parametrizable components. 
Every scenario comes with a set of tests, which run a local heuristic on all agents. Furthermore, we vectorize and port all 9 scenarios from MPE~\cite{lowe2017multi} to VMAS. In this section, we give a brief overview of our new scenarios. For more details (e.g., observation space, reward, etc.) you can find in-depth descriptions in the \href{https://github.com/proroklab/VectorizedMultiAgentParticleSimulator}{VMAS repository}\footref{foot:vmas_url}. 

\vspace{8pt}

\noindent\textbf{Transport (\autoref{fig:transport})}. $N$ agents have to push $M$ packages to a goal. Packages have a customizable mass and shape. Single agents are not able to move a high-mass package by themselves. Cooperation with teammates is thus needed to solve the task.

\vspace{8pt}

\noindent\textbf{Wheel (\autoref{fig:wheel})}. $N$ agents have to collectively rotate a line. The line is anchored to the origin and has a parametrizable mass and length. The team's goal is to bring the line to a desired angular velocity. Lines with a high mass are impossible to push for single agents. Therefore, the team has to organize with agents on both sides to increase and reduce the line's velocity. 

\vspace{8pt}

\noindent\textbf{Balance (\autoref{fig:balance})}. $N$ agents are spawned at the bottom of a world with vertical gravity. A line is spawned on top of them. The agents have to transport a spherical package, positioned randomly on top of the line, to a given goal at the top. The package has a parametrizable mass and the line can rotate.

\vspace{8pt}

\noindent\textbf{Give Way (\autoref{fig:give_way})}. Two agents start in front of each other's goals in a symmetric environment. To solve the task, one agent has to give way to the other by using a narrow space in the middle of the environment.

\vspace{8pt}

\noindent\textbf{Football (\autoref{fig:football})}. A team of $N$ blue agents competes against a team of $M$ red agents to score a goal. By default, red agents are controlled by a heuristic AI, but self-play is also possible. Cooperation among teammates is required to coordinate attacking and defensive maneuvers. Agents need to communicate and assume different behavioural roles in order to solve the task.

\vspace{8pt}

\noindent\textbf{Passage (\autoref{fig:passage})}. 5 agents, starting in a cross formation, have to reproduce the same formation on the other side of a barrier. The barrier has $M$ passages ($M=1$ in the figure). Agents are penalized for colliding amongst each other and with the barrier. This scenario is a generalization of the one considered in~\cite{blumenkamp2021framework}.

\vspace{8pt}

\noindent\textbf{Reverse transport (\autoref{fig:reverse_transport})}. This task is the same as Transport, except only one package is present. Agents are spawned \textit{inside} of it and need to push it to the goal.

\vspace{8pt}

\noindent\textbf{Dispersion (\autoref{fig:dispersion})}. There are $N$ agents and $N$ food particles. Agents start in the same position and need to cooperatively eat all food. Most MARL algorithms cannot solve this task (without communication or observations from other agents) as they are constrained by behavioural homogeneity deriving from parameter sharing. Heterogeneous behaviour is thus needed for each agent to tackle a different food particle.

\vspace{8pt}

\noindent\textbf{Dropout (\autoref{fig:dropout})}. $N$ agents have to collectively reach one goal. To complete the task, it is enough for only one agent to reach the goal. The team receives an energy penalty proportional to the sum of all the agents' controls. Therefore, agents need to organize themselves to send only the closest robot to the goal, saving as much energy as possible. 

\vspace{8pt}

\noindent\textbf{Flocking (\autoref{fig:flocking})}. $N$ agents have to flock around a target without colliding among each other and $M$ obstacles. Flocking has been an important benchmark in multi-robot coordination for years, with first solutions simulating behaviour according to local rules~\cite{reynolds1987flocks}, and more recent work using learning-based approaches~\cite{tolstaya2020flocking}. In contrast to related work, our flocking environment contains static obstacles.

\vspace{8pt}

\noindent\textbf{Discovery (\autoref{fig:discovery})}. $N$ agents have to coordinate to cover $M$ targets as quickly as possible while avoiding collisions. A target is considered covered if $K$ agents have approached a target at a distance of at least $D$. After a target is covered, the $K$ covering agents each receive a reward and the target is re-spawned at a random position. This scenario is a variation of the Stick Pulling Experiment~\cite{ijspeert2001collaboration} and while it can be solved without communication, it has been shown that communication significantly improves performance for $N$ < $M$.

\vspace{8pt}

\noindent\textbf{Waterfall (\autoref{fig:waterfall})}. $N$ agents move from top to bottom through a series of obstacles. This is a testing scenario that can be used to discover VMAS's functionalities.


%% file: sections/mpe_comparison.tex
In this section, we compare the scalability of VMAS and MPE~\cite{lowe2017multi}. Given that we vectorize and port all the MPE scenarios in VMAS, we can compare the two simulators on the same MPE task. The task chosen is ``simple\_spread'', as it contains multiple collidable agents in the same environment. VMAS and MPE use two completely different execution paradigms: VMAS, being vectorized, leverages the Single Instruction Multiple Data (SIMD) paradigm, while MPE uses the Single Instruction Single Data (SISD) paradigm. Therefore, it is sufficient to report the benefits of this paradigm shift on only one task, as the benefits are task-independent.

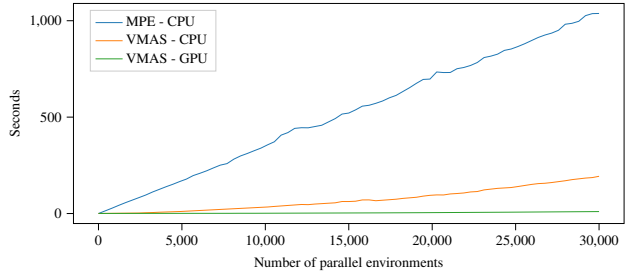
\begin{figure}[h!]
    \centering
    \scalebox{0.6}{%
    \input{figures/mpe_comparison/mpe_comparison}}
    \caption{Comparison of the scalability of VMAS and MPE~\cite{lowe2017multi} in the number of parallel environments. In this plot, we show the execution time of the ``simple\_spread'' scenario for 100 steps. MPE does not support vectorization and thus cannot be run on a GPU.}
    \label{fig:mpe_comparison}
\end{figure}

In \autoref{fig:mpe_comparison}, we can see the growth in execution time with respect to the number of environments stepped in parallel for the two simulators. MPE runs only on the CPU, while VMAS, using PyTorch, runs both on the CPU and on the GPU. In this experiment, we compare the two simulators on an Intel(R) Xeon(R) Gold 6248R CPU @ 3.00GHz and we also run VMAS on an NVIDIA GeForce RTX 2080 Ti. The results show the impact of vectorization on simulation speed. On the CPU, VMAS is up to 5x faster than MPE. On the GPU, the simulation time for VMAS is independent of the number of environments, and runs up to 100$\times$ faster. The same results can be reproduced on different hardware. In the \href{https://github.com/proroklab/VectorizedMultiAgentParticleSimulator}{VMAS's repository}\footref{foot:vmas_url} we provide a script to repeat this experiment.

%% file: figures/mpe_comparison/mpe_comparison.tex
\begin{tikzpicture}

\definecolor{darkgray176}{RGB}{176,176,176}
\definecolor{darkorange25512714}{RGB}{255,127,14}
\definecolor{lightgray204}{RGB}{204,204,204}
\definecolor{steelblue31119180}{RGB}{31,119,180}
\definecolor{green}{RGB}{44,160,44}

\begin{axis}[
scale only axis=true,
width=\linewidth,
height=0.4\linewidth,
legend cell align={left},
legend style={
  fill opacity=0.8,
  draw opacity=1,
  text opacity=1,
  at={(0.03,0.97)},
  anchor=north west,
  draw=lightgray204
},
tick align=outside,
tick pos=left,
title={},
scaled ticks=false, 
x grid style={darkgray176},
xlabel={Number of parallel environments},
xmin=-1498.95, xmax=31499.95,
xtick style={color=black},
y grid style={darkgray176},
ylabel={Seconds},
ymin=-51.8446296572685, ymax=1089.55411685705,
ytick style={color=black}
]
\addplot [semithick, steelblue31119180]
table {%
1 0.0371315479278564
811.783813476562 27.769702911377
1217.17565917969 42.5418014526367
1622.56762695312 56.5375900268555
2433.35131835938 82.8487319946289
2838.7431640625 96.2251586914062
3244.13525390625 111.793746948242
4054.9189453125 139.141738891602
4460.31103515625 151.861343383789
4865.70263671875 165.839370727539
5271.0947265625 178.706115722656
5676.486328125 196.890808105469
6081.87841796875 208.800445556641
6487.2705078125 221.729568481445
6892.662109375 236.750442504883
7298.05419921875 250.950592041016
7703.44580078125 258.885955810547
8108.837890625 281.903930664062
8514.2294921875 298.704986572266
8919.6220703125 311.454803466797
9325.013671875 325.562866210938
9730.4052734375 339.120513916016
10135.796875 356.37646484375
10541.189453125 372.168365478516
10946.5810546875 406.632476806641
11351.97265625 419.812591552734
11757.365234375 441.653930664062
12162.7568359375 445.096435546875
12568.1484375 444.572143554688
13378.9326171875 457.838684082031
14189.7158203125 492.410217285156
14595.1083984375 515.853942871094
15000.5 521.164916992188
15405.8916015625 537.571655273438
15811.2841796875 556.857421875
16216.67578125 561.490966796875
16622.068359375 571.675537109375
17027.458984375 583.693237304688
17432.8515625 600.141357421875
17838.244140625 612.76025390625
18649.02734375 652.895141601562
19054.41796875 675.385620117188
19459.810546875 695.513671875
19865.203125 697.362670898438
20270.59375 734.109313964844
20675.986328125 731.173156738281
21081.37890625 731.542419433594
21486.76953125 751.052856445312
21892.162109375 757.402770996094
22297.5546875 767.898010253906
22702.9453125 783.299621582031
23108.337890625 809
23513.73046875 816.264587402344
23919.12109375 826.696960449219
24324.513671875 846.113037109375
24729.90625 853.262573242188
25135.296875 866.059143066406
25540.689453125 880.532836914062
26351.47265625 913.076965332031
26756.865234375 925.855346679688
27162.255859375 935.866882324219
27567.6484375 950.4296875
27973.041015625 981.765686035156
28378.431640625 986.556518554688
28783.82421875 997.012451171875
29189.216796875 1026.40380859375
29594.607421875 1036.822265625
30000 1037.67236328125
};
\addlegendentry{MPE - CPU}
\addplot [semithick, darkorange25512714]
table {%
1 0.177299976348877
2433.35131835938 2.66015696525574
2838.7431640625 4.14012432098389
3244.13525390625 5.1555871963501
3649.52709960938 6.77866554260254
4865.70263671875 10.3733282089233
10135.796875 33.9152946472168
10541.189453125 36.8092346191406
12162.7568359375 46.9539108276367
12568.1484375 46.0832252502441
12973.541015625 49.1964225769043
14189.7158203125 55.5887069702148
14595.1083984375 62.3101005554199
15000.5 62.238208770752
15405.8916015625 63.8216400146484
15811.2841796875 70.4993515014648
16216.67578125 70.7533569335938
16622.068359375 66.4213714599609
17838.244140625 74.271614074707
18243.634765625 78.513671875
19054.41796875 84.3815383911133
19459.810546875 90.2557373046875
19865.203125 94.3944473266602
20270.59375 96.620964050293
20675.986328125 96.2870635986328
21081.37890625 101.634246826172
21486.76953125 103.821815490723
21892.162109375 106.631042480469
22297.5546875 111.989837646484
22702.9453125 114.092712402344
23108.337890625 122.612533569336
23919.12109375 130.556610107422
24729.90625 134.852554321289
25135.296875 140.009552001953
25946.08203125 150.927124023438
26351.47265625 155.179916381836
26756.865234375 157.005630493164
27162.255859375 160.784942626953
27973.041015625 170.05143737793
28378.431640625 175.843399047852
29189.216796875 183.919906616211
29594.607421875 186.492630004883
30000 193.123809814453
};
\addlegendentry{VMAS - CPU}
\addplot [semithick, green]
table {%
1 0.490082979202271
7879.525390625 1.09798455238342
8485.5654296875 1.35930705070496
9394.6259765625 1.35736846923828
16667.111328125 3.38233280181885
18788.251953125 4.18041896820068
19394.29296875 4.42005729675293
24545.63671875 6.93449306488037
30000 10.1494998931885
};
\addlegendentry{VMAS - GPU}
\end{axis}

\end{tikzpicture}

%% file: sections/experiments.tex
\begin{figure}[ht]
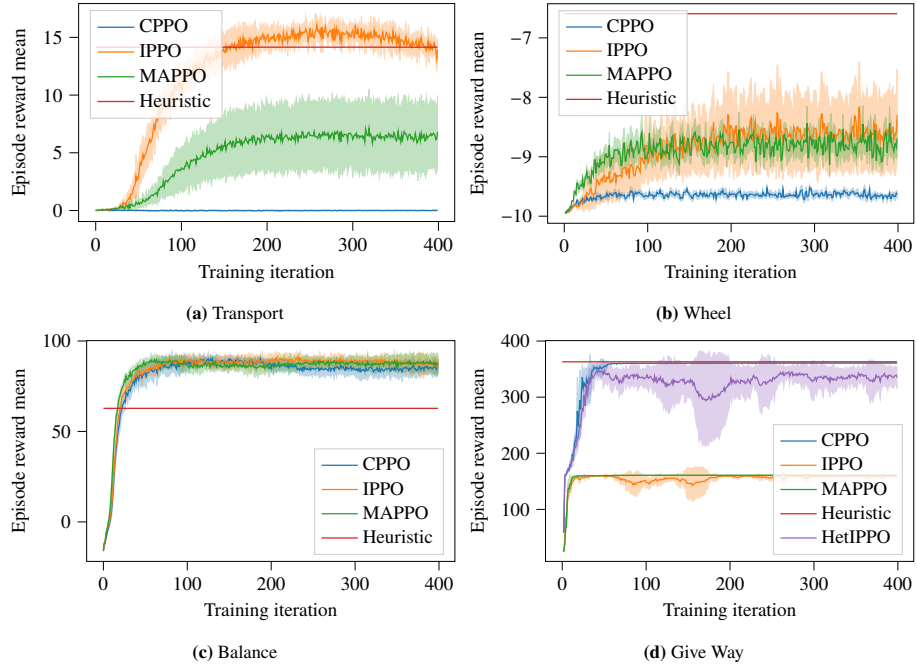

    \centering
    \begin{subfigure}[b]{0.5\linewidth}
        \resizebox{\linewidth}{!}{%
          \input{figures/experiments/transport}}%
           \caption{Transport}
           \label{fig:experiments_transport}
     \end{subfigure}%
    \begin{subfigure}[b]{0.5\linewidth}
            \resizebox{\linewidth}{!}{%
          \input{figures/experiments/wheel}}%
          \caption{Wheel}
          \label{fig:experiments_wheel}
     \end{subfigure}
      \begin{subfigure}[b]{0.5\linewidth}
             \resizebox{\linewidth}{!}{%
          \input{figures/experiments/balance}}%
         \caption{Balance}
          \label{fig:experiments_balance}
     \end{subfigure}%
     \begin{subfigure}[b]{0.5\linewidth}
        \resizebox{\linewidth}{!}{%
          \input{figures/experiments/give_way}}%
         \caption{Give Way}
          \label{fig:experiments_give_way}
     \end{subfigure}
    \caption{Benchmark performance of different PPO-based MARL algorithms in four VMAS scenarios. Experiments are run in RLlib~\cite{liang2018rllib}. Each training iteration is performed over 60,000 environment interactions. We plot the mean and standard deviation of the mean episode reward\footref{foot:reward} over 10 runs with different seeds.}
    \label{fig:experiments}
\end{figure}


We run a set of training experiments to benchmark the performance of MARL algorithms on four VMAS scenarios. Thanks to VMAS's vectorization, we are able to perform a training iteration (comprised of 60,000 environment interactions and deep neural network training) in 25s on average. The runs reported in this section all took under 3 hours to complete. The models compared are all based on Proximal Policy Optimization~\cite{schulman2017proximal}, an actor-critic RL algorithm. The actor is a Deep Neural Network (DNN) which outputs actions given the observations and the critic is a DNN (used only during training) which, given the observations, outputs a value representing the goodness of the current state and action. We refer to the actor and critic as \textit{centralized} when they have access to all the agents' observations and output all the agent's actions/values and we call them \textit{decentralized} when they only map one agent's observations to its action/value. The models compared are:
\begin{itemize}
    \item \textbf{CPPO}: This model uses a centralized critic and actor. It treats the multi-agent problem as a single-agent problem with one super-agent.
    \item \textbf{MAPPO~\cite{yu2021surprising}}: This model uses a centralized critic and a decentralized actor. Therefore, the agents act independently, with local decentralized policies, but are trained with centralized information.
    \item \textbf{IPPO~\cite{de2020independent}}: This model uses a decentralized critic and actor. Every agent learns and acts independently. Model parameters are shared among agents so they can benefit from each other's experiences.
    \item \textbf{HetIPPO}: We customize IPPO to disable parameter sharing, making each agent's model unique.
    \item \textbf{Heuristic}: This is a hand-designed decentralized heuristic different for each task.
\end{itemize}

Experiments are run in RLlib~\cite{liang2018rllib} using the vectorized interface. We run all algorithms for 400 training iterations. Each training iteration is performed over 60,000 environment interactions. We plot the mean and standard deviation of the mean episode reward\footnote{The episode reward mean is the mean of the total rewards of episodes contained in the training iteration\label{foot:reward}} over 10 runs with different seeds. The model used for all critics and actors is a two layer Multi Layer Perceptron (MLP) with hyperbolic tangent activations. A video of the learned policies is available at this \href{https://youtu.be/aaDRYfiesAY}{link}\footref{foot:video}. In the following, we discuss the results for the trained scenarios. 
\vspace{3pt}

\noindent\textbf{Transport (\autoref{fig:experiments_transport})}. In the Transport environment, only IPPO is able to learn the optimal policy. This is because the other models, which have centralized components, have an input space consisting of the concatenation of all the agents' observations. Consequently, centralized architectures fail to generalize in environments requiring a high initial exploration like this one, where there is a high variance in possible joint states (and therefore there is a low probability that a similar state will be encountered).

\vspace{3pt}

\noindent\textbf{Wheel (\autoref{fig:experiments_wheel})}. The Wheel environment proved to be a hard task for MARL algorithms. Here, all models were not able to solve the task and performed worse than the heuristic. 

\vspace{3pt}

\noindent\textbf{Balance (\autoref{fig:experiments_balance})}. In Balance, all models were able to solve the task and outperform the heuristic. However, this is largely due to the use of a big observation space containing global information. The task can be made arbitrarily harder by removing part of the observation space and thus increasing partial observability.

\vspace{3pt}

\noindent\textbf{Give Way (\autoref{fig:experiments_give_way})}. In the Give Way scenario, it is shown that only algorithms able to develop heterogeneous agent behaviour can solve the environment. In fact, IPPO and MAPPO, which use parameter sharing and decentralized actors, fail this scenario. On the other hand, it is shown that the scenario can be solved either through a centralized actor (CPPO) or by disabling parameter sharing and allowing agent policies to be heterogeneous (HetIPPO).

The experimental results confirm that VMAS proposes a selection of scenarios which prove challenging in orthogonal ways for current state-of-the-art MARL algorithms. We show that there exists no one-fits-all solution and that our scenarios can provide a valuable benchmark for new MARL algorithms.  In addition, vectorization enables faster training, which is key to a wider adoption of multi-agent learning in the robotics community.

%% file: sections/conclusion.tex
In this work, we introduced \href{https://github.com/proroklab/VectorizedMultiAgentParticleSimulator}{VMAS}, an open-source vectorized simulator for multi-robot learning. VMAS uses PyTorch and is composed of a core vectorized 2D physics simulator and a set of multi-robot scenarios, which encode hard collective robotic tasks. The focus of this framework is to act as a platform for MARL benchmarking. Therefore, to incentivize contributions from the community, we made implementing new scenarios as simple and modular as possible. We showed the computational benefits of vectorization with up to 30,000 parallel simulations executed in under 10s on a GPU. We benchmarked the performance of MARL algorithms on our scenarios. During our training experiments, we were able to collect 60,000 environment steps and perform a training iteration in under 25s. Experiments also showed how VMAS scenarios prove difficult in orthogonal ways for state-of-the-art MARL algorithms. In the future, we plan to extend the features of VMAS to widen its adoption, continuing to implement new scenarios and benchmarks. We are also interested in modularizing the physics engine, enabling users to swap vectorized engines with different fidelities and computational demands.